\documentclass[
]{ceurart}

\usepackage{listings}
\usepackage{adjustbox}
\usepackage{tabularx}

\begin{document}

\copyrightyear{2024}
\copyrightclause{Copyright for this paper by its authors.
  Use permitted under Creative Commons License Attribution 4.0
  International (CC BY 4.0).}

\conference{CHR 2024: Computational Humanities Research Conference, December 4–6, 2024, Aarhus, Denmark}

\title{Handwriting Recognition in Historical Documents with Multimodal LLM}

\tnotemark[1]

\author{}
\begin{abstract}
  There is an immense quantity of historical and cultural documentation that exists only as handwritten manuscripts. At the same time, performing OCR across scripts and different handwriting styles has proven to be an enormously difficult problem relative to the process of digitizing print. While recent Transformer based models have achieved relatively strong performance, they rely heavily on manually transcribed training data and have difficulty generalizing across writers. Multimodal LLM, such as GPT-4v and Gemini, have demonstrated effectiveness in performing OCR and computer vision tasks with few shot prompting. In this paper, I evaluate the accuracy of handwritten document transcriptions generated by Gemini against the current state of the art Transformer based methods. 
\end{abstract}


\maketitle

\section{Introduction}

The vast majority of historical documents exist only in manuscript form. Correspondence, personal notes, ledgers, and other unpublished documents provide an enormous source of data that is completely inaccessible to computational text analysis methods. Even for traditional humanists, the lack of keyword search and indexing makes the research process significantly less efficient. Research in these areas requires painstaking and difficult close reading, or the mobilization of huge numbers of volunteer transcribers as in \cite{causer2018making}.  These barriers prove prohibitive in most cases and limit the scope and scale of potential questions. Unlike the easy and open access to digitized print documents provided by HathiTrust and Internet Archive, databases of handwritten historical documents are nearly nonexistent. These absences in digital availability of sources warp the types of questions scholars are able to pursue, even in traditional humanistic pursuits where travel to physical sources could prove prohibitively expensive. Moreover, recognition and understanding of historical handwriting is often difficult even for subject experts. 

While OCR on print has achieved extremely high accuracy (outside of non-Latin types, low quality scanning, and unconventional page formatting) since the early 2000s, high  accuracy handwriting OCR was essentially impossible until the adoption of image based convolutional neural networks in studies such as (\cite{dutta2018improving}).  These methods are not only computationally intensive, but generally require pretraining or fine tuning on annotated subsets of the specific corpus, creating nearly insurmountable programming barriers for non technically proficient researchers or those without annotated data.

Thus, the development of automated digitization of handwritten documents is an important tool for archives and cultural institutions to open the way for a wide range of new historical research projects, using both conventional close reading and computational approaches.

\section{Literature Review}
Previous attempts to perform the handwriting task, detailed in surveys like \cite{baldominos2019survey} and \cite{babu2019character}, have generally focused on machine vision models, specifically CNN. However, without a underlying language model, the task was difficult, as many letterforms are inconsistent and indistinguishable, especially in cursive script. An underlying language model was necessary to provide support and predictions in cases of visual ambiguity. 
Initial studies, like  \cite{dutta2018improving}, utilized a combined CNN - BiLSTM architecture. The current state of the art (\cite{dolfing2020whole}, \cite{parres2023fine},) use a combined vision transformer and text transformer model based on the TrOCR model \cite{li2023trocr}. This combines a vision transformer encoder with a Roberta based text transformer decoder. TrOCR is trained on synthetic handwriting data and then finetuned for specific domain applications. TrOCR represents the state of the art for general handwriting recognition outside of the historical or multilingual context. Both of these methods rely on extensive preprocessing to align text, detect line breaks, and addressing marginalia, again posing significant barriers to users without technical expertise, along with the aforementioned issues with fine tuning on labelled data. 

Multimodal LLMs, such as GPT-4v \cite{achiam2023gpt}, have displayed promising early ability to recognize text and tabular structure in images. Although architecture and training details are unclear, these models seem to directly take image data along with text prompts as input into a text centric Transformer model.  Papers such as \cite{yang2023dawn} and \cite{shi2023exploring} have shown, though limited exploratory examples, that GPT-4 is able to effectively transcribe English handwriting with high accuracy as well as adapt to tabular structure and non aligned text, two problems which previous studies required extensive preprocessing to address. However, these papers have not conducted a comprehensive study on the accuracy of transcription on a robust evaluation dataset. Evaluation of handwriting recognition quality in LLM also focuses on contemporary examples of handwriting, making it difficult to draw conclusions about performance in the historical context. 

\section{Data}
A wide range of multilingual corpora have been published as an evaluation set for this problem. \cite{SANCHEZ2019122} describes the image and gold standard transcription pairs presented at the ICDAR conference between 2014 and 2017, including a 1,200 page excerpt from Jeremy Bentham's papers in Early Modern English, a 450 page set of Early Modern German, and a 150 page dataset of cursive modern Latin. As supplemental sets, \cite{dolfing2020scribblelens} provides 200 pages of 17th Century Dutch documents. An additional Spanish language dataset, RODRIGO \cite{serrano2010rodrigo} provided line images for training the SoTA models. These documents were chosen because they have already been used to train and evaluate state of the art models and they represent well structured gold standard labelled data.

\begin{figure}[hbt!]
    \centering
    \includegraphics[width=0.8\linewidth]{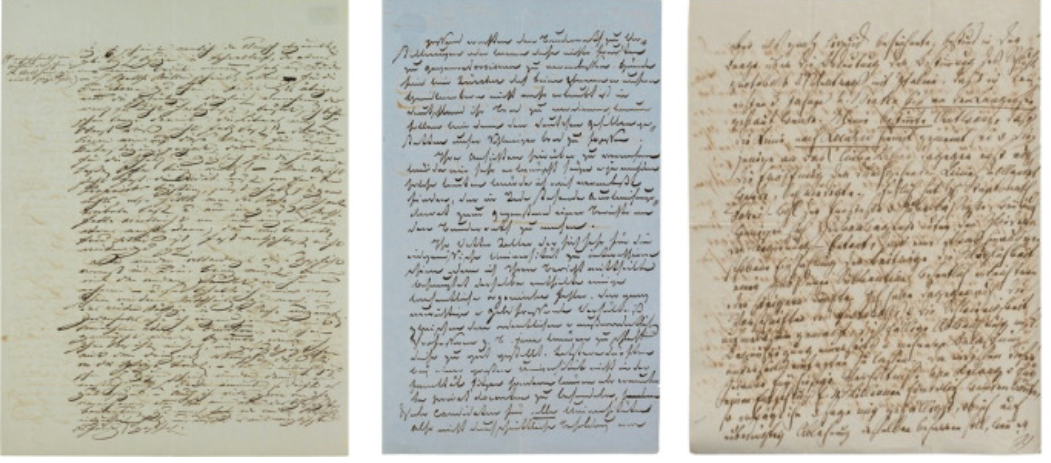}
    \caption{Example data from \cite{SANCHEZ2019122}}
    
\end{figure}
\begin{figure}[hbt!]
    \centering
    \includegraphics[width=0.8\linewidth]{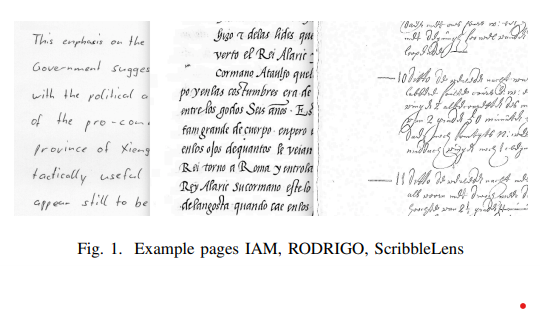}
    \caption{Example data from \cite{dolfing2020scribblelens}}
    
\end{figure}
\begin{figure}[hbt!]
    \centering
    \includegraphics[width=\linewidth]{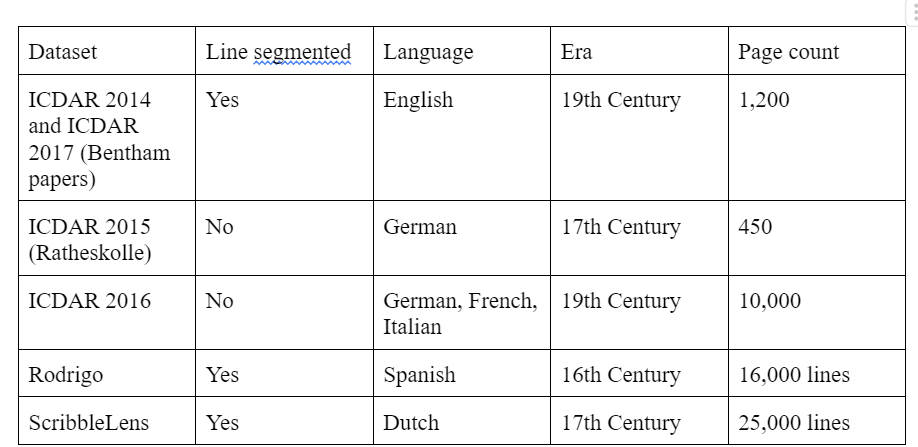}
    \caption{Metadata Characteristics}
    \label{fig:enter-label}
\end{figure}

\section{Methods}

\subsection{Reproduction}
I reimplement a CNN-BiLSTM and a fine-tuned TrOCR architecture based on the SOTA papers. This will be applied to the different letters to evaluate word and character level accuracy to reproduce the results. 

For the CNN, I implemented the code at \cite{chammas2018handwriting} with default hyperparameter settings and roughly 500 epochs of training. 

For TrOCR, I fine tuned using the HuggingFace pipeline and TrOCR-Base-Stage-1 with default settings. 

Both pre-existing methods were trained or fine tuned using GPU compute on Google Colab. 

The data was split by language for training and evaluation. The ICDAR 2014 and 2017 Bentham dataset provided 1,200 English samples, ICDAR 2015 provided 450 German documents, and ICDAR 2016 provided 6,000 German and 4,000 French documents. These three languages were selected because they represented the largest collections of languages from specific time periods and corpora available across pre-published data. For each language, I randomly sampled 30 and 500 annotated pages for the training/fine tuning. 

\subsection{Gemini}
The closed source nature of Gemini and other corporate LLMs makes it impossible to fine tune or train with additional data, so the model is applied directly without any modification to fit the specific problems. I selected Gemini, specifically the gemini-pro-vision model, primarily for cost reasons, as the API access is currently free. I evaluate three different prompting strategies. One with just a request to transcribe with no revisions (simple), one providing contextual info on the language and time period of the input image (background), and a final approach where I provide contextual info and ask the model to correct spelling and grammar errors. 
The simple prompt: "Transcribe the following document exactly with no modifications:". 
Background prompt: "Transcribe the following document exactly. It is from the {X} century and in {Y} language:"
Correction Prompt: "Transcribe the following document. It is from the {X} century and in {Y} language. Correct any spelling or grammar errors:"

On the initial evaluations, we observe no significant difference between the three prompting strategies. All accuracies were within 1-2 percentage points for each Gemini prompting strategy. This suggests that there is limited impact of providing additional contextual information through prompting beyond the image alone.

It is currently impossible to provide few shot image examples to Gemini because prompts can only contain one image, but as the feature is added I will evaluate further. 

\section{Results}
\subsection{Model Performance}
\begin{table}[ht]
\centering
\caption{Character error rate comparison across approaches}

\begin{tabularx}{\linewidth}{ r X X X X X X X}
\toprule
            Language& CNN-BiLSTM 30 samples& CNN-BiLSTM 500 samples&TrOCR base& TrOCR 30 samples finetune & TrOCR 500 sample finetune& Gemini \\ \midrule
English& 40.1\%& \textbf{6.8\%}                    & 62\%                   & 31\%                    & 7.5\%&34\% 
\\\
French& 43.1\%& \textbf{9.8\%}                    & 81\%                   & 43\%                    & 10.7\%&56\% \\
German& 42.1\%& 10.1\%                   & 92\%                   & 53\%                    &\textbf{9.2\%}&74\% \\ \bottomrule
\end{tabularx}
\end{table}
\begin{figure}[hbt!]
    \centering
    \includegraphics[width=0.8\linewidth]{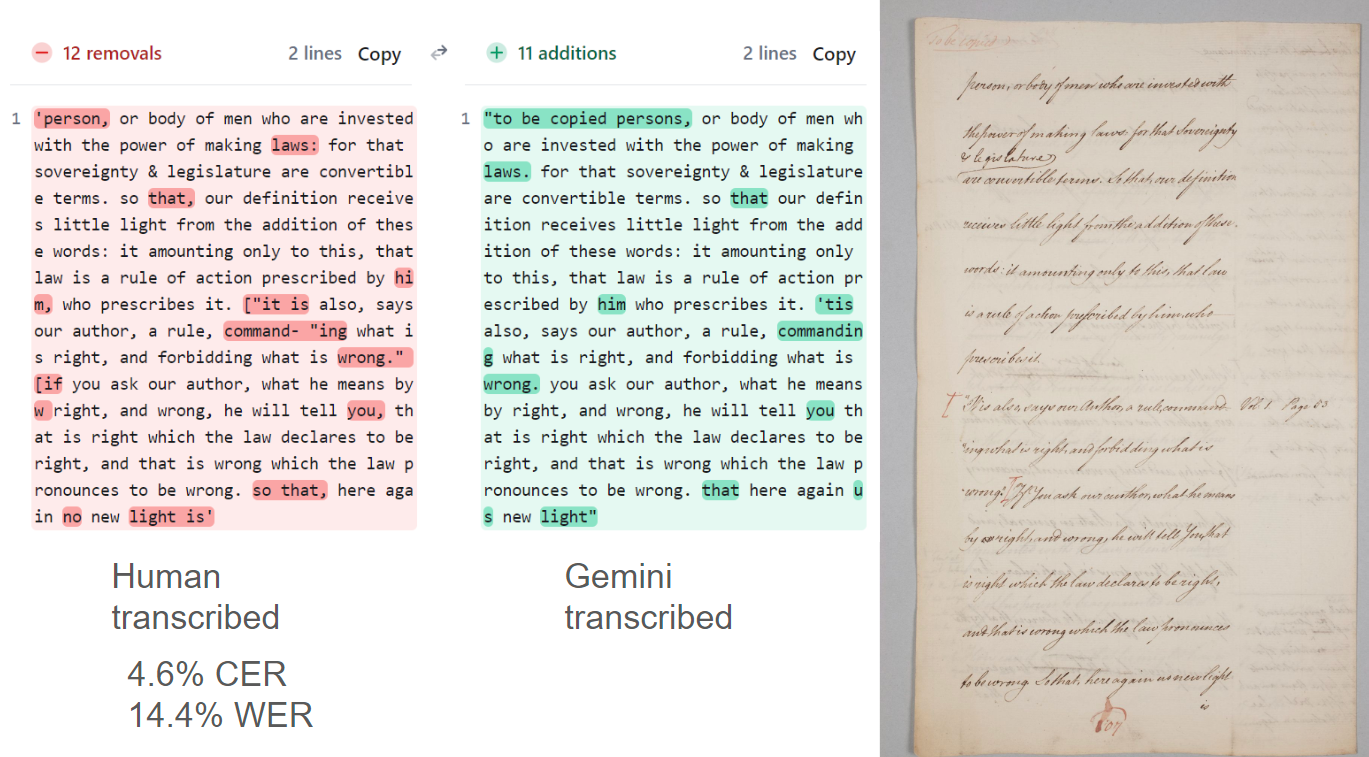}
    \caption{Gemini performance on Bentham document}
    
\end{figure}
Figure 4 shows the Gemini output with the prompt: "transcribe the following document. It is in 17th century English by Jeremy Bentham. Correct any spelling errors." and a non pre-processed image provided. We can see in figure 4 that the English language performance of zero shot Gemini transcription is comparable to the SOTA models finetuned on the Bentham corpus. The majority of errors are omissions or incorrect interpretation of punctuation marks (e.g. interpreting a colon as a period or missing brackets added in a different ink color).
\begin{figure}[hbt!]
    \centering
    \includegraphics[width=0.8\linewidth]{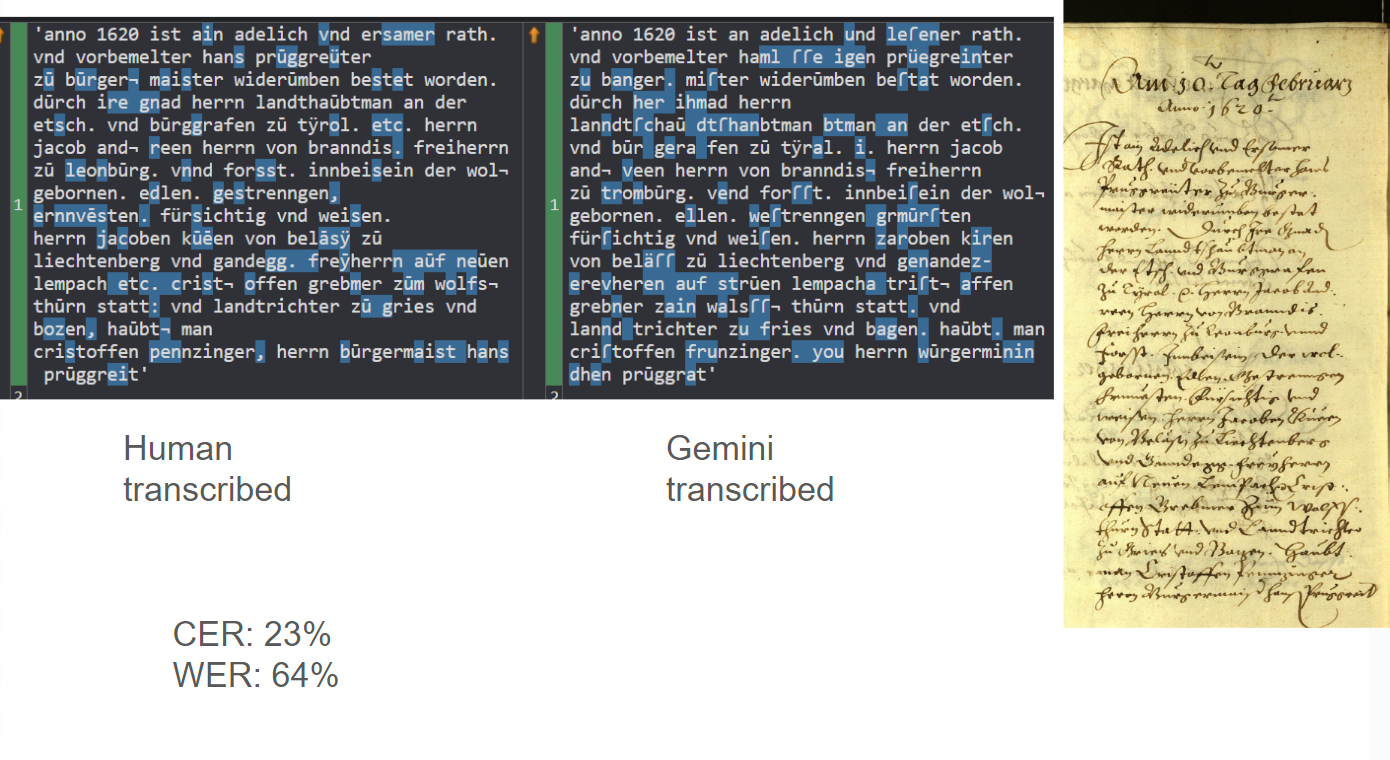}
    \caption{Gemini performance on German document}
    
\end{figure}
Figure 5 shows the Gemini output with the prompt: "transcribe the following document. It is in 17th century German." The performance is dramatically worse, and also worse than the SOTA models performance finetuned on 17th century German. This suggests that the English dominated training corpus of Gemini, especially images of handwriting from which it derives emergent handwriting recognition abilities may be impacting performance on certain tasks. 

While the overall performance is much worse than fine tuned TrOCR or the CNN, especially for non-English languages, there are some promising signs. First, for English and to a lesser extent French and German, the model displays a broad range of performance (Figure 6), with a large number of perfectly transcribed samples, and some high error samples in German resulting from difficult to read text suggesting that the emergent capabilities are present, but more efforts to prevent hallucinations such as lowering temperature or prompting techniques are necessary. When the model fails, it is generally not because of incorrect transcriptions; model errors generally resulted from text generation, rather than letterform recognition. In cases with extremely high inaccuracy, the model generally produced text completely unrelated to the underlying image as a result of hallucination or other model errors (Figure 7). Such hallucinations result from the model's text generation process, and account for the majority of high error documents, which skews the performance of Gemini. Applying models trained to detect common hallucinations may also reduce the error rate from these occurrences.

\begin{figure}[hbt!]
    \centering
    \includegraphics[width=\linewidth]{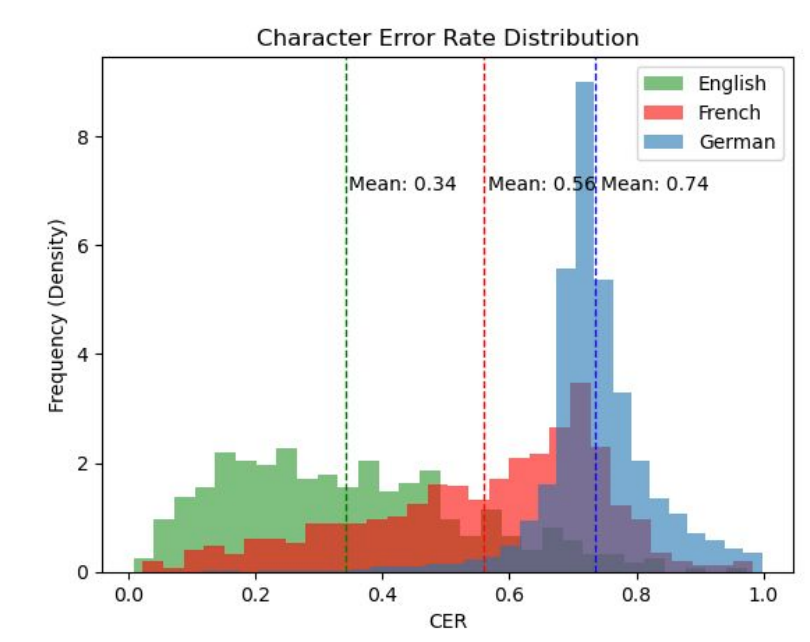}
    \caption{Error Rate Distribution of Gemini}
    
\end{figure}

\begin{figure}[hbt!]
    \centering
    \includegraphics[width=1\linewidth]{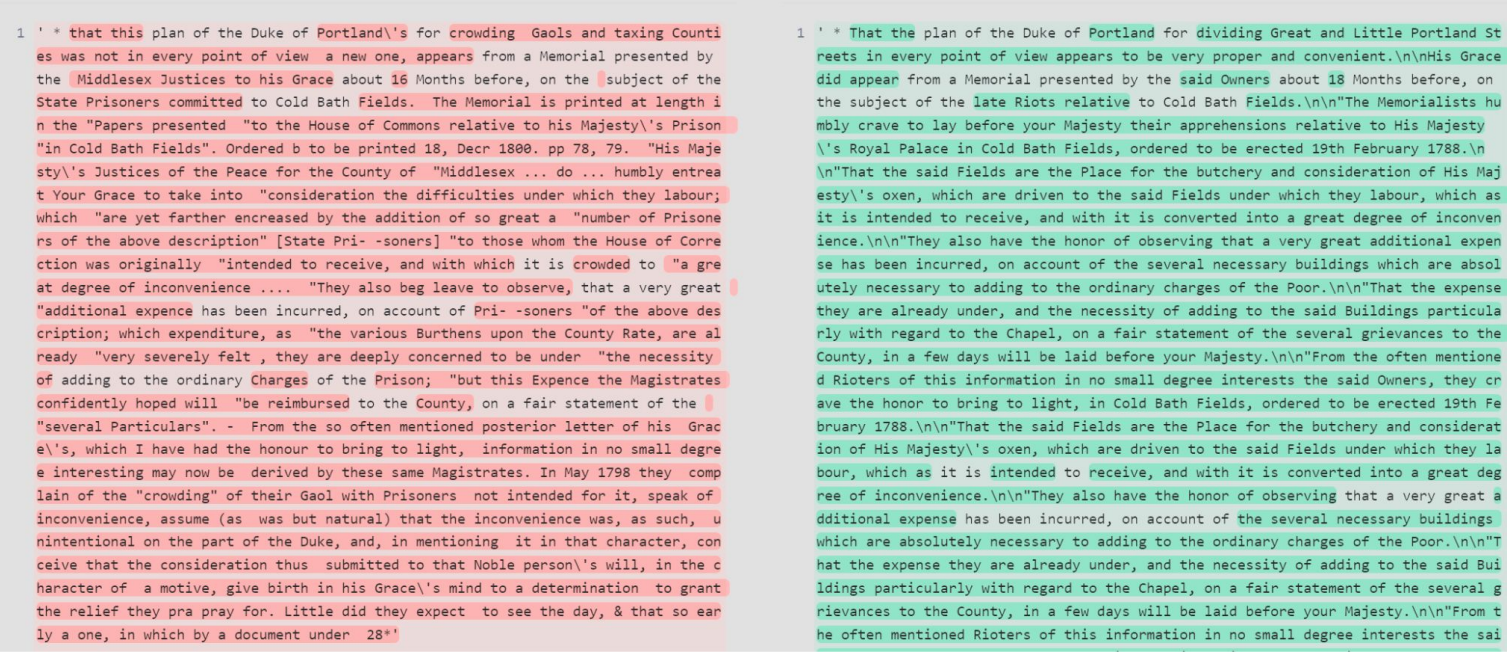}
    \caption{Hallucinations in Gemini}
    \label{fig:enter-label}
\end{figure}
\section{Discussion}
\subsection{Recommendations}

There are no clear advantages to the state of the art trained models unless a sufficiently large in-domain annotated dataset is provided from the same language and time period. With 30 training samples, the English language performance of both published models is not noticeably superior to zero shot base Gemini. We observe far superior performance with a large number of annotated fine tuning samples, but it is important to remember that this high accuracy is only for a specific corpus, and cannot be easily transferred across languages, time periods, or even specific authors. 

Biases in representation of non English languages in the Gemini training dataset makes it much weaker for non-English languages. For these applications, it is still necessary to use trained neural models. 

Finetuning with TrOCR, especially with the unique affordances of HuggingFace for simplified training and model dissemination, appears to offer more cost effective and efficient performance relative to RCNNs. The fine tuning process is both simpler to implement with fewer potential pitfalls versus the raw python code of the RCNN model. At both small and large fine tuning thresholds, TrOCR achieves comparable or superior performance to the RCNN model with a much more streamlined training process. 

However, for projects without access to extensive computational resources and/or access to labor for manual training set annotation, Gemini and other multimodal LLMs offers a potentially simpler alternative. Gemini is more accurate at transcribing English language sources than state of the art models with no training or small training sets. On non-English languages, trained models are still necessary. 

\begin{figure}[h!]
    \centering
    \includegraphics[width=\linewidth]{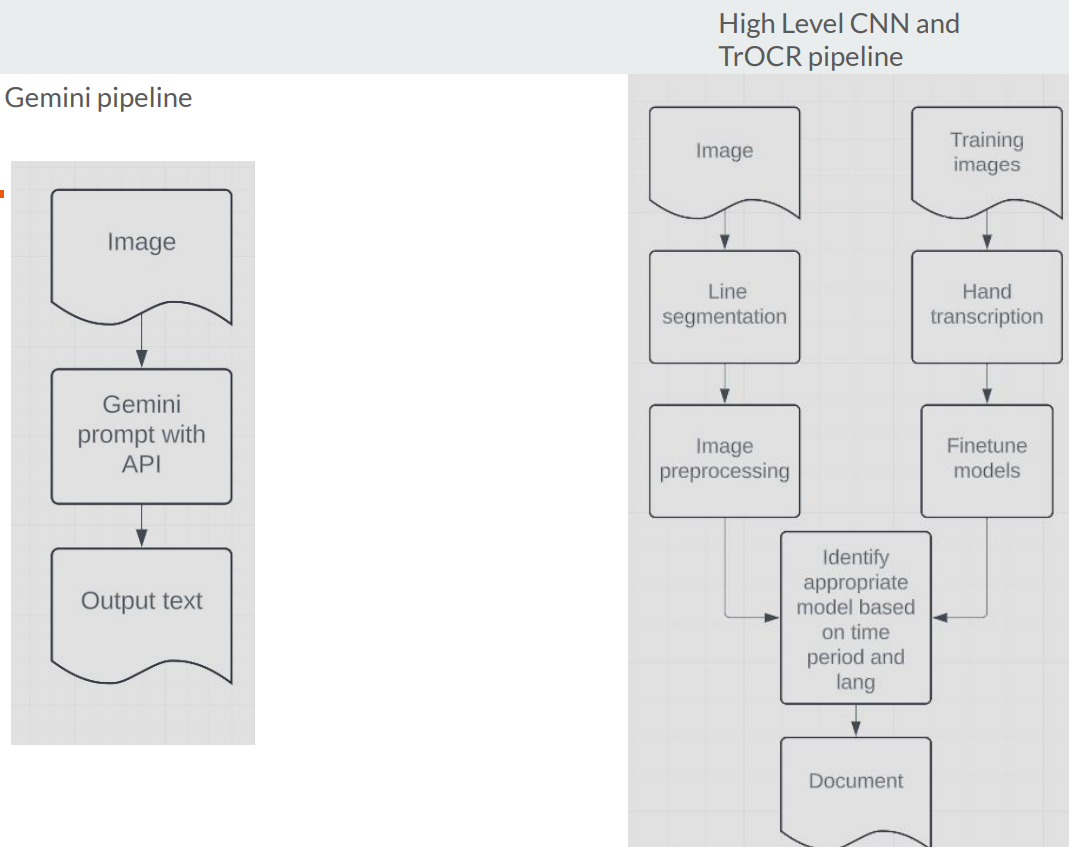}
    \caption{Gemini vs SoTA workflows}
    \label{fig:enter-label}
\end{figure}
It is important to also consider the infrastructure necessary to support digitization approaches, and not just the raw accuracy rate.  Gemini also allows for programatically simple and free access (Figure 8), in contrast to previous methods, which rely on access to GPU resources, previous experience with fine tuning or training neural language models, and more complicated feature engineering of samples, including line segmentation, contrast, and greyscale conversion. While utilities like HuggingFace have greatly simplified this process, the CNN's older and less documented code was far more complex to replicate. For cultural institutions and non-technical researchers, employing Gemini or other cloud hosted pretrained LLMs may greatly simply workflows and reduce technical barriers to entry. 

For future work on this project, I plan to also evaluate the cross corpus effectiveness of TrOCR, especially on texts from different periods and languages outside the training data. This will ideally evaluate the transfer learning capabilities of the model and provide a more fair comparision to the zero shot example of Gemini. I also plan to experiment with prompting techniques or model settings to reduce the hallucination prevalence of Gemini and other LLMs.

\begin{acknowledgments}

\end{acknowledgments}

\bibliography{bibliography}

\section{Online Resources}
Github and datasets removed for anonymization purposes

\end{document}